# PotatoPestNet: A CTInceptionV3-RS-Based Neural Network for Accurate Identification of Potato Pests

Md. Simul Hasan Talukder[1]
Bangladesh Atomic Energy Regulatory Authority, Bangladesh;
E-mail: simulhasantalukder@gmail.com

Rejwan Bin Sulaiman[2]
Northumbria University, UK;
E-mail: rejwan.binsulaiman@gmail.com

Mohammad Raziuddin Chowdhury[3]
Jahangirnagar University, Bangladesh;
E-mail: razichy3@gmail.com

Musarrat Saberin Nipun[4]
Brunel University London UK;
email: musarrat.nipun@brunel.ac.uk

Taminul Islam[5]
Department of Computer Science and Engineering, Daffodil International University, Bangladesh;
Email: taminul15-11116@diu.edu.bd

**Corresponding Author:** Rejwan Bin Sulaiman; Northumbria University, UK; E-mail: rejwan.binsulaiman@gmail.com

**Abstract:** Potatoes are the third-largest food crop globally, but their production frequently encounters difficulties because of aggressive pest infestations. Early classification those potato pests plays an important role in the detection and prevention of their notorious attack. The aim of this study is to investigate the various types and characteristics of these pests and propose an efficient PotatoPestNet AI-based automatic potato pest identification system. To accomplish this, we curated a reliable dataset consisting of eight types of potato pests. We leveraged the power of transfer learning by employing five customized, pre-trained transfer learning models: CMobileNetV2, CNASLargeNet, CXception, CDenseNet201, and CInceptionV3, in proposing a robust PotatoPestNet model to accurately classify potato pests. To improve the models' performance, we applied various augmentation techniques, incorporated a global average pooling layer, and implemented proper regularization methods. To further enhance the performance of the models, we utilized random search (RS) optimization for hyperparameter tuning. This optimization technique played a significant role in fine-tuning the models and achieving improved performance. We evaluated the models both visually and quantitatively, utilizing different evaluation metrics. The robustness of the models in handling imbalanced datasets was assessed using the Receiver Operating Characteristic (ROC) curve. Among the models, the Customized Tuned Inception V3 (CTInceptionV3) model, optimized through random search, demonstrated outstanding performance. It achieved the highest accuracy (91%), precision (91%), recall (91%), and F1-score (91%), showcasing its superior ability to accurately identify and classify potato pests.

## 1. Introduction

Food security around the world is a serious concern. According to the report of the director general of FAO [1], 50% of food demand will increase by 2050. To face the challenge, the production of crops must be increased. Among the crops, potato is the third largest food crop after rice and wheat across the world [1]. It is a globally indispensable crop with considerable economic, nutritional, and food security value [2–3]. In 2019, the value of global potato exports was estimated to be around $6.2 billion USD, with the largest exporters being France, the Netherlands, and Germany. It is grown in more than 125 countries, of which China and India are the largest producers [4]. A statistic on the amount of potato production and area of land based on cultivation all over the world is shown in Figures 1 and 2, respectively [4].

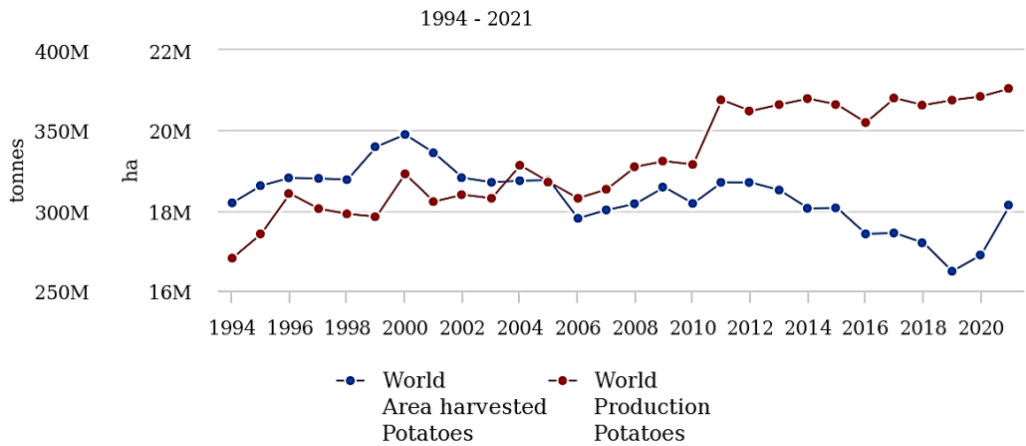

**Figure 1.** Potato production/ yield quantities in the world [Source: FAOSTAT, May 6, 2023].

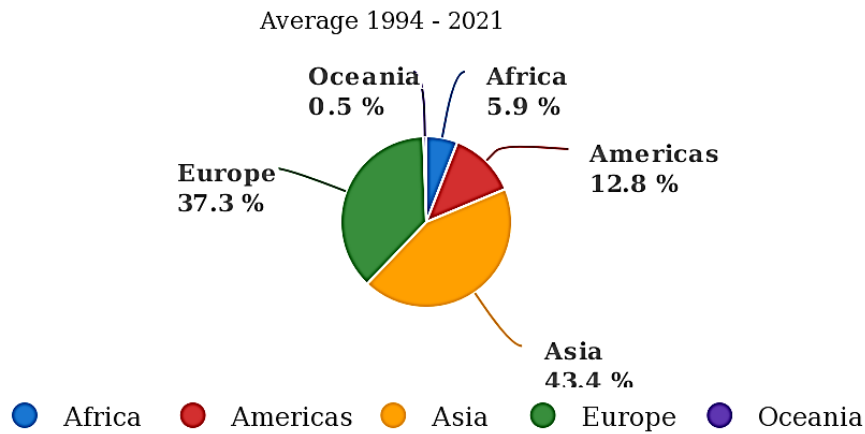

**Figure 2.** Production share of potatoes by region [Source: FAOSTAT, May 6, 2023].

However, potato production is severely hampered by pests and diseases, which cause major output losses and diminish crop quality. Nineteen types of potato pests were recorded in a study of Bangladesh [5]. Among them, eight types of pests are more harmful. They are Agrotis ipsilon (Hufnagel), Amrasca devastans (Distant), Aphis gossypii Glover, Bemisia tabaci (Gennadius), Brachytrypes portentosus Lichtenstein, Epilachna vigintioctopunctata (Fabricius), Myzus persicae (Sulzer), and Phthorimaea operculella (Zeller).

For effective management and control, detection and identification of pests at an early stage are crucial. At the outset of this study, it was required to determine farmers' knowledge of potato pests and their methods for treating diseases caused by potato or potato leaf pests. Okonya et al. [6] conducted research in Uganda, where they surveyed agricultural households to determine the farmers' awareness of potato pests. This study was conducted in Uganda's six districts in August and September 2013. The survey question was designed so that they could identify the most susceptible pests based on interview responses. The study allowed the researchers to identify the most severe and moderately severe pests. Only 5% of the 204 farmers were able to accurately answer questions about insect pests, according to the results of the study. That means the farmers are not well-versed in pest identification. They need to go to the nearest agriculturist, which is time consuming and a hassle. To solve this issue, artificial intelligence (AI) is one of the best ways to diagnose and identify potato pests. With the development of technology, machine learning algorithms have been created to identify and classify pests, making the procedure more efficient and precise [7]. Algorithms for machine learning can analyze huge volumes of data rapidly and precisely [8], enabling farmers to make informed decisions regarding crop management and resource allocation. This can lead to better efficiency in the use of resources, including water, fertilizer, and pesticides, which can ultimately result in larger yields and lower costs. Using those algorithms, it is possible to detect and categorize crop illnesses and pests, enabling early intervention and the reduction of crop losses. This can improve the crop's quality and reduce waste. Also, machine learning algorithms can aid farmers in adopting more sustainable agricultural methods [9], such as precision agriculture and integrated pest management. This can decrease the environmental impact of farming and increase long-term sustainability.

This study investigated the application of machine learning algorithms in potato pest recognition and analyzed the maximum level of precision. Using five customized tuned (CT), and pre-trained machine learning techniques, which are CTDenseNet201, CTMobileNetV2, CTNASLargeNet, CTXception, and CTInceptionV3, on our prepared dataset, this paper proposes a PotatoPestNet model that has the highest efficiency and robustness in detecting and identifying various forms of potato crop pests. The findings of this study will assist potato farmers and researchers in the development of efficient and effective pest management measures, ultimately resulting in increased potato yields and enhanced crop quality. The main contribution of this study can be summarized as follows:

- Preparation of Potato Pest dataset.
- Increasing the size of dataset and reducing using different augmentation techniques.
- Modifying five pre-trained model.
- Replacing fully connected layer and dropout layer to reduce further level of overfitting.
- Application of random search technique for hyper parameters tuning of the models.
- Comparing the performance of the models to find the best model.
- Proposing PotatoPestNet by fine tuning the CTInceptionV3-RS based transfer learning model.
- Performance evaluation of PotatoPestNet model and comparing with the earliest research.

Section 2 of this paper has detailed other related research effort. Section 3 contains the method and materials. Section 4 depicts the experimental setup. Sections 5, 6, 7, and 8 exhibit the result analysis, discussion, limitation, and conclusion, respectively.

## 2. Literature Review

The food demand of a growing global population is making agriculture more important than ever. Scientists are working on various aspects at once to find ways to increase agricultural production and guarantee a constant supply of fresh crops. Potatoes are one of the most extensively consumed crops in the world, which means that their production is necessary in almost every region. As a result, scientists are always looking for new hardware and software methods in agriculture that may help to make larger production.

Five years ago, the researchers did AI research in the agricultural field, detecting different plant diseases and nutrient deficiencies by leaf image processing. Many of the traditional machine learning algorithms like SVM, random forest, decision tree, k-means clustering, and so on were more popular [10–11]. Some researchers tried to use ANN techniques in leaf disease classification [12]. Due to the huge number of training parameters and large computation time, ANN is replaced by CNN. However, many comparative studies have revealed that the traditional models are less effective in plant disease classification. R. Sujatha et al. [13] showed a comparative study between the traditional ML model and deep learning models for citrus leaf disease detection. VGG16, VGG19, and InceptionV3 were the most prominent in classification accuracy. Similarly, Sunil S. Harakannanavar [14] also summarized the effectiveness of CNN machine learning techniques in detecting tomato leaf disease. Further, C. Jackulin et al. [15] found that the deep learning models performed better in plant disease diagnosis. But the problem is the requirement for a large dataset. To avoid the problem of a limited dataset and to get better performance, the researchers started to apply the transfer learning concept.

Since our study was related to potato pest identification, we tried to focus on the studies on AI in potatoes. Many research studies have been conducted on potato leaf diseases recognition. Rabbia Mahum et. al. [16] proposed an efficient DenseNet201 by adding an extra transition layer to classify five types of potato leaf diseases. A reweighted cross entropy loss function was used to make the model more robust. Regularization was used to reduce the overfitting. The efficient DenseNet201 models had an accuracy of 97.2%. The authors did not focus on data preprocessing widely here. But Alok Kumar et. al. [17] did it and proposed a hierarchical deep learning convolutional neural network (HDLCNN) model to identify the potato leaf disease. The median filtering method was applied to remove noise in dataset preprocessing. The author introduced the intuitionistic fuzzy local binary pattern (IFLBP) to extract features and applied HDLCNN to classify the potato leaf diseases. The accuracy of HDLCNN was approximately 4% greater than that of VGG-INCEP, Deep CNN, Random Forest methods (RF), and other spiking neural networks (SNN) methods. Transfer learning may not always perform the best. Ali Arshaghi et. al. [18] showed it. He designed a convolutional neural network (CNN) to detect five classes of potato diseases and compared it with Alexnet, Googlenet, VGG, R-CNN, and transfer learning models. The CNN model outperformed the other models and contributed 99% accuracy. Many scientists tried to modify pre-train models in potato disease classification for better performance. Mosleh Hmoud Al-Adhaileh et. al. [19] customized a convolutional neural network (CNN) to detect early blight, late blight, and healthy leaves of potatoes. It outperformed the existing work with 99% accuracy. The author did not use any tuning techniques. Alberta Odamea Anim-Ayeko et al. [20]

proposed the ResNet-9 model for detecting the blight disease state of potato and tomato leaf images, where hyperparameter optimization was used in their work that gave 99.25% test accuracy.

Most of the studies are on the diagnosis of potato leaf diseases. We have tried to review some of the previous work on crop pest identification as well as potato pest diagnosis. Mei-Ling Huang et al. [21] utilized deep learning for feature extraction of the tomato pest's dataset and classified it using conventional machine learning models named discriminant analysis (DA), support vector machine, and k-nearest neighbor method (KNN). Bayesian optimization was used for hyperparameter tuning. VGG16 after the augmentation exhibited 94.95% accuracy. But ResNet50 with the discriminant analysis model achieved 97.12% accuracy. Nurul Nabilah Ahmad Loti et al. [22] used six traditional feature-based approaches and six deep learning feature-based approaches to extract the significant features, which were then classified by SVM, RF, and ANN for the chili plant diseases and pest infestation. The outcome of this research was that the deep learning feature-based approaches performed better, and the SVM classifier achieved the best accuracy of 92.10%. That means the studies used a hybrid concept of traditional machine learning and deep learning models. In consideration of time as well as performance, Cheng et al. [23] employed a convolutional neural network (CNN) model and Xie's research database to classify images of 10 agricultural pests. They utilized the visual geometry group 16 (VGG16) model, which achieved an accuracy rate of 95.33%. Additionally, they developed a pest detection system using a faster R-CNN model that was capable of processing images with complex backgrounds and accurately detecting and tracking pests, even those with protective coloring. In just pest identification, Md. Sakib Ullah Sourav et al. [24] suggested a transfer learning model named VGG19 to identify the most vital four types of jute pests. His proposed model provided 95.86% accuracy. He did not check the response of another pre-trained models.

From the literature review, it is clearly found that no study has been accomplished on potato pest identification by using machine learning. Most of the research works are executed in the disease diagnosis of potatoes by leaf image processing. Previously, though traditional ML models were preferred, they are being replaced by CNN models, transfer learning models, hybrids of them, and many other approaches. Some of the research was on pest identification in different plants. In our study, we successfully prepared a dataset of potato pests and proposed a CTInceptionV3-RS-Based PotatoPestNet model to identify potato pest efficiently and robustly.

## 3. Materials and Methodology

Automatic early diagnosis of potato pests is a significant development in the field of agriculture that has been accomplished through this research. We have proposed a robust and efficient PotatoPestNet machine learning model to detect potato pests. The work flow and the entire process are presented below.

### 3.1. Dataset Preparation

With the advent of artificial intelligence in agriculture, a new field of research has emerged to enhance its application and performance. But preparing a dataset of sufficient size is the ultimate obstacle. In this study, we did not get any publicly available datasets on potato pests. There was also no recognized institution or source from which we could collect the data. That's why our team prepared the dataset by web scraping using the open-source Python library named downloader. download (). There are 19 pests for potato cultivation [5]. But due to the limitations of images in Google Images, online portals, newspapers, and publicly shared samples, we have focused on the eight most commonly deleterious types of potato pests. The name of the pest and number of finally collected images are listed in Table 1 and shown in Figure 3. The collected images were checked manually by an agriculturist before being placed in the proper class. In the last steps, data pre-processing operations such as enhancing the contrast, cropping to remove unnecessary portions, background removal were applied. Thus, ultimately, we prepared the whole dataset, which has a total of 495 images in eight classes. The dataset was split into training, testing, and validation data with a ratio of 70:15:15, as shown in Table 2.

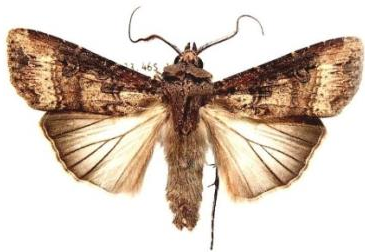

(a) Agrotis ipsilon (Hufnagel)

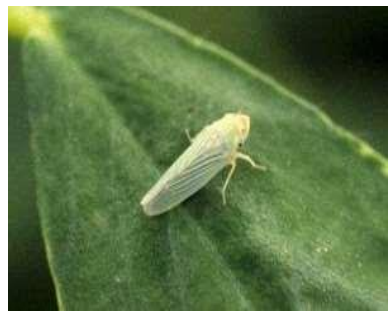

(b) Amrasca devastans (Distant)

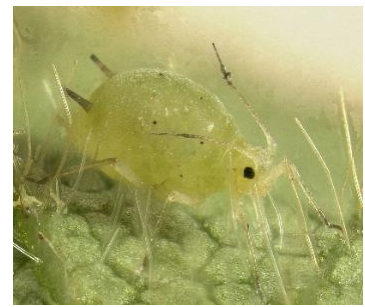

(c) Aphis gossypii Glover

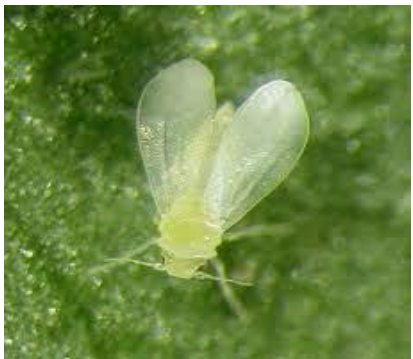

(d) Bemisia tabaci (Gennadius)

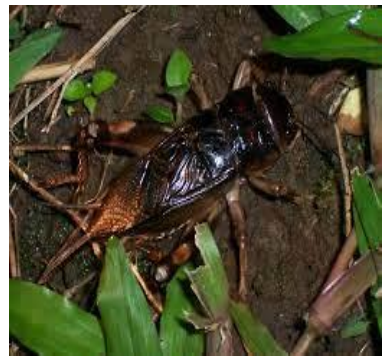

(e) Brachytrypes portentosus Lichtenstein

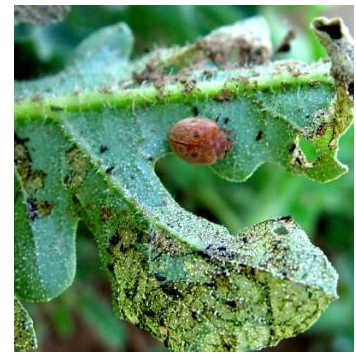

(f) Epilachna vigintioctopunctata (Fabricius)

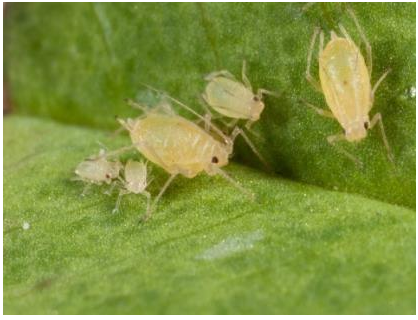
(g) Myzus persicae (Sulzer)

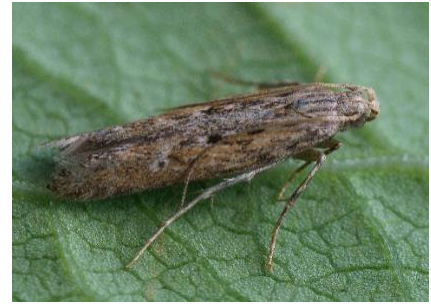
(h) Phthorimaea operculella (Zeller)

Figure 3. Visual presentation of our dataset.

**Table 1.** Name of the potato pests and corresponding number of images.

| Name of potato pest | Number of images |
|---|---|
| Agrotis ipsilon (Hufnagel) | 139 |
| Amrasca devastans (Distant) | 62 |
| Aphis gossypii Glover | 37 |
| Bemisia tabaci (Gennadius) | 35 |
| Brachytrypes portentosus Lichtenstein | 35 |
| Epilachna vigintioctopunctata (Fabricius) | 70 |
| Myzus persicae (Sulzer) | 75 |
| Phthorimaea operculella (Zeller) | 42 |
|  | Total= 495 |

**Table 2**. Training, testing and validation dataset.

| Training data | Testing data | Validation data |
|---|---|---|
| 342 | 81 | 71 |

## 3.2. Data Augmentation

In many fields, like medical, agriculture, and so on, the collection of large numbers of images is very arduous. In image classification using machine learning, the larger the dataset, the greater the effectiveness of the model [25]. Data augmentation is a technique by which the number of images can be increased. It greatly contributes to reducing the overfitting of models [26–27]. According to many studies, this augmentation technique not only reduces overfitting but also improves the accuracy of the model and deals with regularization problems [28]. In our work, we used the augmentation method due to the smaller number of images in the dataset. During the training phase, we utilized six different augmentation techniques on the training dataset. The augmentation procedures used in our work are summarized in algorithm 1. Moreover, the number of iterations in our work was one. After augmentation, the number of training images was 2268.

| | **Algorithm 1. Augmentation procedures** |
|---|---|
| 1 | Input: Training dataset |
| |       n: number of iterations |
| 2 | Output: Augmented training dataset |
| 3 | For i=0 to n |
| 4 |     rotation_range=30 |
| 5 |     zoom_range=0.2 |
| 6 |     `width_shift_range=0.2 |
| 7 |     height_shift_range=0.2 |
| 8 |     vertical_flip=True |
| 9 |     horizontal_flip=True |
| 10 | End |
| 11 | Return the augmented training dataset |
| 12 | End of the Algorithm` |

## 3.3. Transfer learning

For the effective design of a new CNN model, a large dataset is always required. But there is no massive, labeled dataset of potato pests. Transfer learning (TL) is one of the solutions to this challenge. It is a process by which the knowledge gained in solving a problem is used to solve similar types of problems [29]. The concept of transfer learning is depicted in Figure 4. The TL models trained on a large ImageNet dataset with 1000 classes can be used for feature extraction on the small dataset. The It reduces the training time, overfitting the model, and enhances the model's accuracy [30]. In our work, we have used five prominent pre-trained models, such as DenseNet201, MobileNetV2, NASNetLarge, Xception, and InceptionV3. The trainable, non-trainable, and total parameters are listed in Table 3. It is obvious that the trainable parameters are reduced, thereby reducing the computational time.

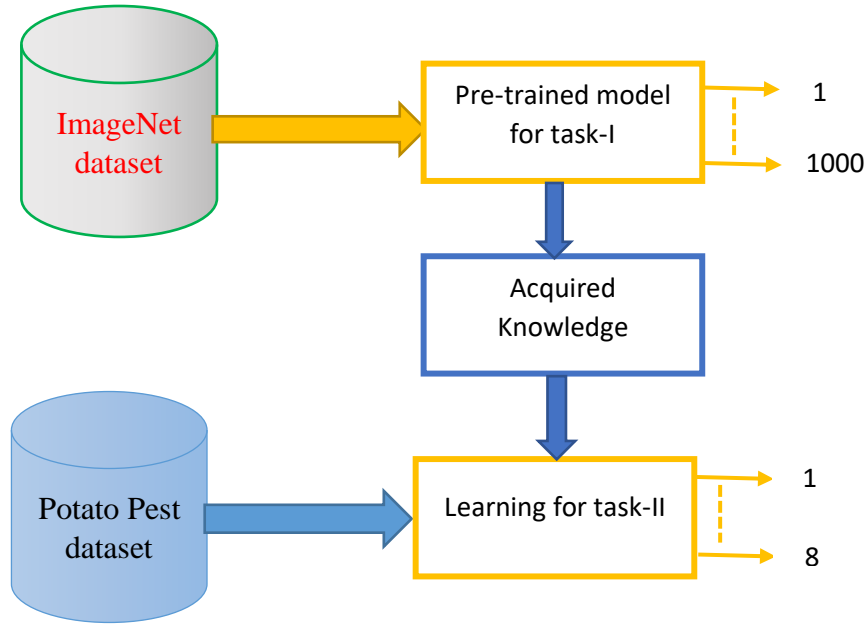

**Figure 4.** Transfer Learning approach.

**Table 3.** Trainable and non-trainable parameters of customized TL models.

| Name of Models | Total Parameters | Non-trainable parameters | Trainable parameters |
|---|---|---|---|
| TCMobileNetV2 | 2,268,232 | 2,257,984 | 10,248 |
| TCNASNetLarge | 84,949,082 | 84,916,818 | 32,264 |
| TCXception | 20,877,872 | 20,862,480 | 16,392 |
| TCDenseNet201 | 18,337,352 | 18,321,984 | 15,368 |
| TCInceptionV3 (PotatoPestNet) | 21,819,176 | 21,802,784 | 16,392 |

### 3.4. Random Search

Random search algorithm (RSA) is a method to find the optimal set of hyperparameters for machine learning models [31]. The parameters that are set before the training of the models are called hyperparameters. Learning rate, dropout rate, optimizer, and so on are examples of hyperparameters. In the random search technique, the values of hyperparameters are specified in a definite range. The algorithm picks randomly a set of parameters and analyzes the model with this parameter. It continues to the end and finally chooses the best-performing parameters [32]. If $\Theta$ is the search space, $\theta_i$ are the hyperparameters (i=1, 2, 3…. n), n is the set of hyperparameters and $\Theta^*$ are the best performing parameter of a function $f(\theta)$, the mathematically it can be expressed like the equation 1.

$$\Theta^* = \text{argmax/min} (f(\theta_i)) \quad \ldots\ldots\ldots\ldots (1)$$

Where max/min denotes the maximization or minimization of the function.

The whole procedures are summarized in algorithm 2.

| | **Algorithm 2. Random Search Technique for hyper parameter tunning** |
|---|---|
| 1 | `Inputs:<br>$f$: Function that need to optimize<br>$\Theta$: Search space of hyperparameters<br>n: Number of iterations |
| 2 | Outputs:<br>$\Theta^*$ Optimal set of hyperparameters |
| 3 | Let $\Theta^*$ = void |
| 4 | For i =1 to n do |
| 5 |    Sample a set of hyperparameters $\Theta\_i$ uniformly at random from $\Theta$ |
| 6 |    Evaluate the performance metric $f(\Theta\_i)$ for $\Theta\_i$ |
| 7 |    If $\theta^*$ is null or $f(\Theta\_i) < f(\Theta^*)$ then set $\Theta^* = \Theta\_i$ |
| 8 | End |
| 9 | Return $\Theta^*$ |
| 10 | End of the Algorithm |

## 3.5. Proposed framework

The main intention of the proposed robust PotatoPestNet model is to identify potato pests automatically, reducing time and human effort, and thereby improving the treatment and production of potatoes. In our study, the first challenge was the limited size of the data. By using the augmentation technique, we have increased the size of our dataset. But still, it is small for the CNN model. To avoid the problem of the small size of the dataset, we have used the TL concept. Five well known pre-trained models, such as DenseNet201, MobileNetV2, NASNetLarge, Xception, and InceptionV3, are used in the study. The TL models are already trained on a large ImageNet dataset with 1000 classes. The models are not directly used in our work. The layers of feature extraction in the models are kept unchanged, but the classification layers are changed, as shown in Figure 5. The fully connected layer is replaced by a global average pooling layer to minimize the overfitting of the model. In addition to regularization, a drop out layer was also applied to reduce model overfitting at another level, and finally, a dense layer with the SoftMax activation function was added to classify the potato pest images.

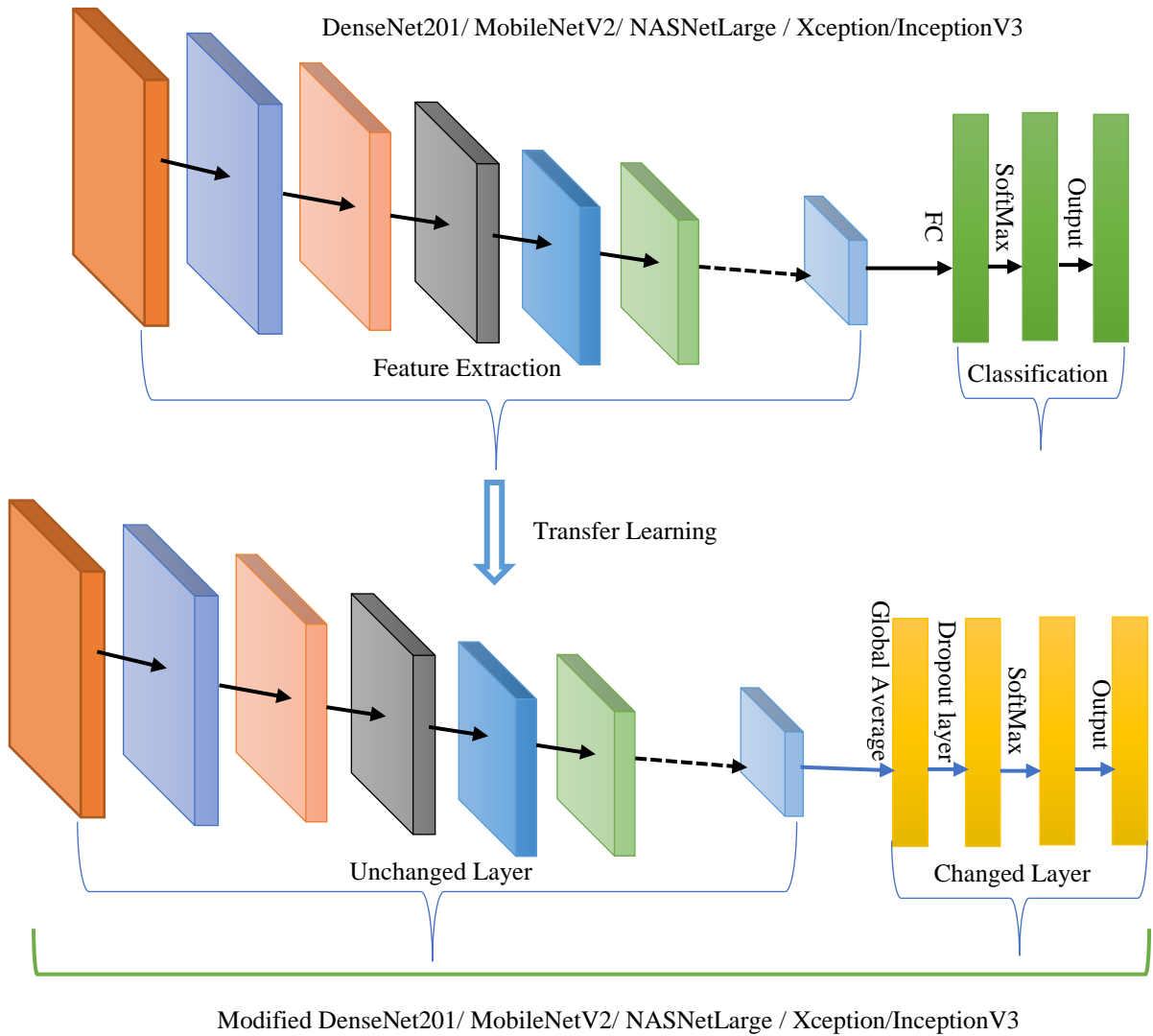

**Figure 5.** Customized pre-trained models.

For the training of the models, we have used categorical cross-entropy loss function to measure the differences between the predicted probability distribution and the true probability distribution. The other parameters are listed in Table 4. Since our dataset is small, there is a high probability of overfitting and bad performance. That's why we tuned the hyperparameters using random search techniques. We selected three hyperparameters for random search: optimizer, dropout rate, and learning rate. The random search method is already described in algorithm 1. The search space is given in Table 5. After getting the tuned parameters, the models are named hyper-parameter tuned modified models, which are pre-trained models that are trained with training data and validation data. Then the models are tested with our test dataset. The CTInceptionV3 model outperforms the others, and it is chosen and named after PotatoPestNet. The whole procedure is depicted in Figure 6.

**Table 4.** The synopsis of the parameters used in the models.

| Performance Measures | Customized DenseNet201 |
|---|---|
| No of epochs | 50 |
| Batch Size | 16 |
| Image Size | 224*224 |
| Activation function | SoftMax |
| Loss | Categorical cross-entropy |

**Table 5.** Random search space.

| Parameters | Values |
|---|---|
| Optimizers | ['adam', 'rmsprop', 'sgd'] |
| Learning rate | [1e-1, 1e-2, 1e-3, 1e-4, 1e-5] |
| Dropout rate | [0.2, 0.3, 0.4, 0.5] |

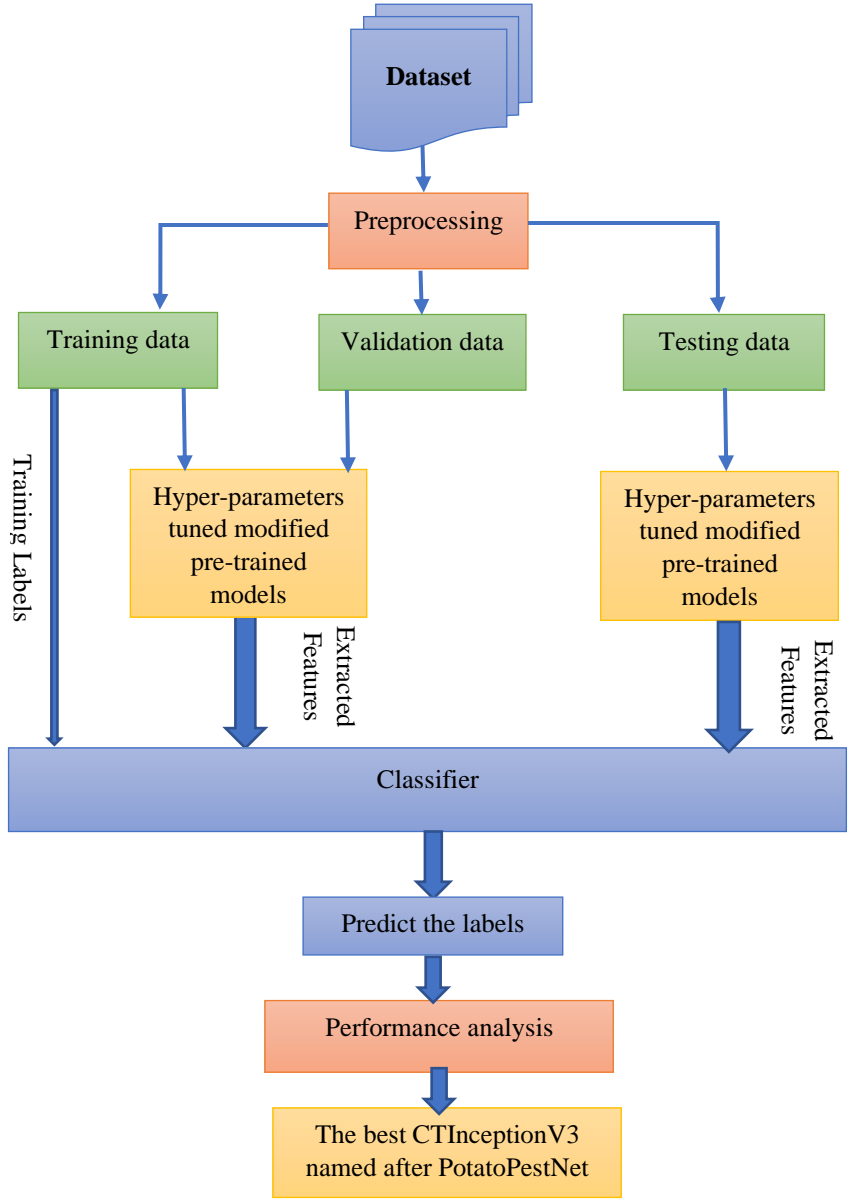

**Figure 6.** Proposed PotatoPestNet framework for potato pest identification.

### 3.6. Model Evaluation

Model evaluation is the way to assess the performance of a model. It is a measure of how well the model is able to classify the new data. There are qualitative and quantitative methods to evaluate. In our work, both are used to assess the models. The performance metrics that are used, are precision, recall, f1-score, accuracy, ROC, and confusion matrix.

**Precision:** It is the proportion of true positive sample to the all-positive prediction. It is given by-

$$\text{Precision} = \frac{TP}{TP+FP} * 100 \qquad (2)$$

**Recall:** The proportion of true positive instance among the actual positive instances is call recall and given by-

$$\text{Recall} = \frac{TP}{TP+FN} * 100 \quad (3)$$

**F1-score:** It is the harmonic mean of precision and recall. It can be defined by-

$$\text{F1 score} = 2 * \left(\frac{Precision*Recall}{Precision+Recall}\right) * 100 \quad (4)$$

**Accuracy:** It indicates how many instances are classified perfectly by the model. It can be calculated by the following equation.

$$\text{Accuracy} = \frac{TP+TN}{TP+TN+FP+FN} \quad (5)$$

**ROC**: ROC (Receiver Operating Characteristic) is a graphical representation of the performance of a binary classification model, which shows the trade-off between the true positive rate (TPR) and the false positive rate (FPR) as the discrimination threshold of the model is varied.

**Confusion Matrix:** A confusion matrix is a table that summarizes the performance of a machine learning model on a classification task, by comparing the predicted class labels with the actual class labels in the test dataset.

**N.B.** TP denotes the true positive; TN is the true negative; FP denotes the false positive; FN denotes the false negative.

## 4. Experimental Setup

To implement and evaluate the proposed model, we used Collaboratory, which runs on a virtual machine (VR). In our session, VR used multiple CPU cores and was GPU-backed. The available disk storage was around 100 GB, and the RAM was 12 GB. In our experiment, we utilized the Python 3.7 programming language along with specific versions of various libraries. The versions used were Keras 2.3.1 and TensorFlow 2.0. These libraries provided us with essential tools and functions for building and training our machine learning models. For data visualization purposes, we employed the Matplotlib and Seaborn libraries. These libraries enabled us to create insightful plots, charts, and graphs to visualize our data, model performance, and other relevant information. Internet with 100 Mbps speed was connected without interruption.

## 5. Result analysis

The objective of our study was to develop an artificial intelligence expert system for accurately identifying potato pests. To achieve this, we employed five prominent pre-trained models, namely CTInceptionV3, CTXception, CTNASNetLarge, CTMobileNetV2, and CTDenseNet201. These models were customized and fine-tuned using different combinations of hyperparameters obtained through the random search (RS) technique. The RS technique involved conducting 10 trials, each consisting of 20 epochs, to explore various hyperparameter configurations. The goal was to identify the optimal combination of hyperparameters that maximized the validation accuracy. The results of the RS technique, including the best hyperparameter configurations, are presented in Tables 6, 7, 8, 9, and 10. The most effective combination of hyperparameters for each pre-trained model is highlighted, bolted, and underlined in the last row of the respective table. To summarize the tuned parameters, Table 11 provides an overview of the selected hyperparameters for each model. The CDenseNet201, CMobileNetV2, and CNASNetLarge models achieved better accuracy when trained with the Adam optimizer, employing a learning rate of 0.0001 and dropout rates of 0.3, 0.4, and 0.3, respectively. On the other hand, the CXception and CInceptionV3 models performed optimally with the SGD optimizer, utilizing a learning rate of 0.1 and dropout rate of 0.4, respectively.

**Table 6.** The outcome of CDenseNet201 with different hyperparameters combinations.

| Model Name | Combination of the hyper parameters | | | Validation Accuracy (%) |
|---|---|---|---|---|
| | **Dropout rate** | **Learning rate** | **Optimizer** | |
| CDenseNet201 | 0.4 | 0.00001 | SGD | 14.08 |
| | 0.5 | 0.01 | Adam | 28.16 |
| | 0.4 | 0.1 | Adam | 28.16 |
| | 0.4 | 0.0001 | SGD | 28.21 |
| | 0.3 | 0.001 | Adam | 30.98 |
| | 0.5 | 0.0001 | SGD | 42.25 |
| | 0.3 | 0.001 | RMSprop | 43.66 |
| | 0.3 | 0.00001 | Adam | 85.91 |
| | 0.4 | 0.0001 | Adam | 92.95 |
| | **_0.3_** | **_0.0001_** | **_Adam_** | **_95.77_** |

**Table 7.** The outcome of CMobileNetV2 with different hyperparameters combinations.

| Model Name | Combination of the hyper parameters | | | Validation Accuracy (%) |
|---|---|---|---|---|
| | **Dropout rate** | **Learning rate** | **Optimizer** | |
| CMobileNetV2 | 0.4 | 0.1 | RMSprop | 28.16 |
| | 0.3 | 0.01 | RMSprop | 28.16 |
| | 0.4 | 0.01 | RMSprop | 28.16 |
| | 0.2 | 0.01 | RMSprop | 28.16 |
| | 0.4 | 0.0001 | SGD | 33.80 |
| | 0.4 | 0.001 | Adam | 35.21 |
| | 0.5 | 0.00001 | RMSprop | 54.92 |
| | 0.4 | 0.001 | SGD | 64.78 |
| | 0.5 | 0.0001 | RMSprop | 69.01 |
| | **_0.4_** | **_0.0001_** | **_Adam_** | **_80.28_** |

**Table 8.** The outcome of CDenseNet201 with different hyperparameters combinations.

| | Combination of the hyper parameters | | | Validation Accuracy (%) |
|---|---|---|---|---|
| | **Dropout rate** | **Learning rate** | **Optimizer** | |
| CNASNetLarge | 0.5 | 0.0001 | SGD | 9.85 |
| | 0.2 | 0.00001 | SGD | 9.85 |
| | 0.2 | 0.0001 | SGD | 22.53 |
| | 0.5 | 0.1 | RMSprop | 28.16 |
| | 0.3 | 0.1 | RMSprop | 28.16 |
| | 0.2 | 0.1 | RMSprop | 29.57 |
| | 0.2 | 0.001 | RMSprop | 36.61 |
| | 0.2 | 0.00001 | RMSprop | 67.6 |
| | 0.3 | 0.1 | SGD | 80.28 |
| | **0.3** | **0.0001** | **Adam** | **84.5** |

**Table 9.** The outcome of CDenseNet201 with different hyperparameters combinations.

| Model Name | Combination of the hyper parameters | | | Validation Accuracy (%) |
|---|---|---|---|---|
| | **Dropout rate** | **Learning rate** | **Optimizer** | |
| CInceptionV3 | 0.4 | 0.1 | Adam | 28.16 |
| | 0.2 | 0.01 | Adam | 28.61 |
| | 0.4 | 0.01 | RMSprop | 32.39 |
| | 0.3 | 0.001 | SGD | 40.84 |
| | 0.5 | 0.00001 | RMSprop | 53.52 |
| | 0.5 | 0.00001 | Adam | 61.97 |
| | 0.4 | 0.00001 | Adam | 67.6 |
| | 0.5 | 0.01 | SGD | 76.05 |
| | 0.2 | 0.01 | SGD | 80.28 |
| | **0.4** | **0.1** | **SGD** | **91.54** |

**Table 10.** The outcome of CDenseNet201 with different hyperparameters combinations.

| Model Name | Combination of the hyper parameters | | | Validation Accuracy (%) |
|---|---|---|---|---|
| | **Dropout rate** | **Learning rate** | **Optimizer** | |
| CXception | 0.4 | 0.1 | Adam | 28.16 |
| | 0.2 | 0.01 | Adam | 28.61 |
| | 0.4 | 0.01 | RMSprop | 32.39 |
| | 0.3 | 0.001 | SGD | 40.84 |
| | 0.5 | 0.00001 | RMSprop | 53.52 |
| | 0.5 | 0.00001 | Adam | 61.97 |
| | 0.4 | 0.00001 | Adam | 67.6 |
| | 0.5 | 0.01 | SGD | 76.05 |
| | 0.2 | 0.01 | SGD | 80.28 |
| | **0.4** | **0.1** | **SGD** | **91.54** |

Table 11. Summary of the tuned hyper parameters.

| Performance Measures | Customized DenseNet201 | Customized MobileNetV2 | Customized NASNetLarge | Customized Xception | Customized InceptionV3 |
|---|---|---|---|---|---|
| Optimizers | Adam | Adam | Adam | SGD | SGD |
| Learning rate | 0.0001 | 0.0001 | 0.0001 | 0.1 | 0.1 |
| Dropout rate | 0.3 | 0.4 | 0.3 | 0.4 | 0.4 |

Following the determination of the best hyperparameters, we individually trained each model using our training and validation datasets. The analysis of the trained models is presented in Figure 7, which displays the validation loss, validation accuracy, training loss, and training accuracy curves for all five pre-trained models. As the number of epochs increased, we observed a consistent decrease in the loss curves and an increase in the accuracy curves. This behavior indicates that the models progressively learned from the data and converged to a well-fitted state. Notably, CTInceptionV3 (PotatoPestNet) exhibited the highest accuracy and the most precise fitting of the accuracy and loss curves.

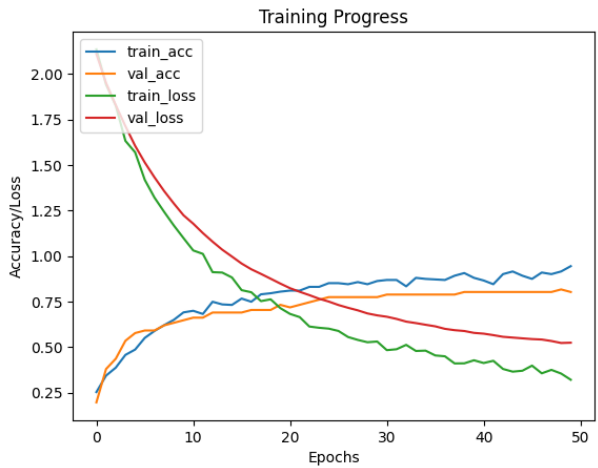
(a) Customized Tuned MobileNetV2.

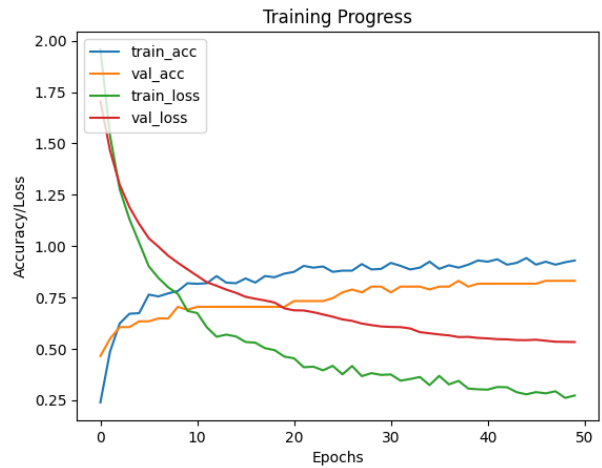
(b) Customized Tuned NASNetLarge.

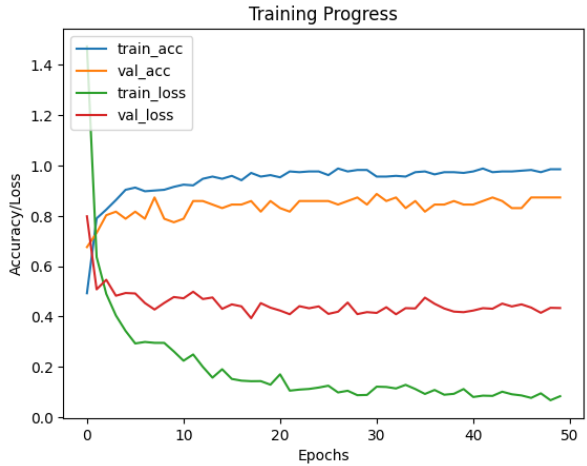
(c) Customized Tuned Xception.

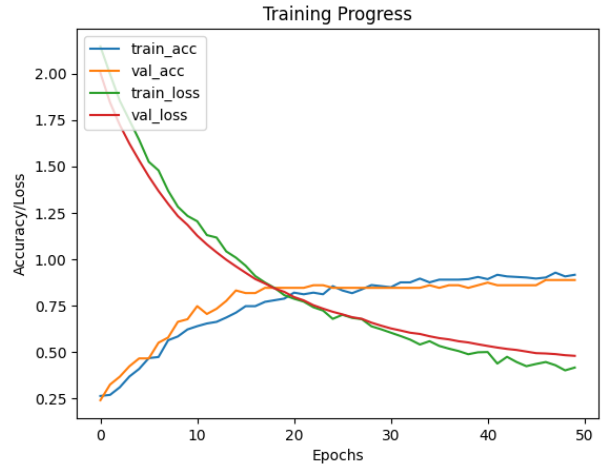
(d) Customized Tuned DenseNet201.

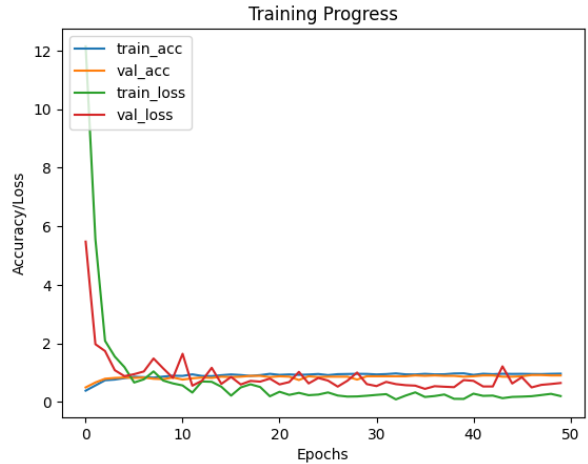
(e) Customized Tunned InceptionV3 (PotatoPestNet).

**Figure 7**. Accuracy and loss curve.

After the training phase, we assessed the models' performance using an independent testing dataset. The evaluation involved visually analyzing the confusion matrices, as depicted in Figure 8. These matrices provide a detailed overview of the models' accurate classifications and misclassifications. While each model displayed different strengths and weaknesses across various classes, all models performed well overall. CTInceptionV3 (PotatoPestNet) demonstrated superior classification performance with only 7 misclassifications, compared to 8, 11, 13, and 14 misclassifications for CTDenseNet201, CTXception, CTNASNetLarge, and CTMobileNetV2, respectively.

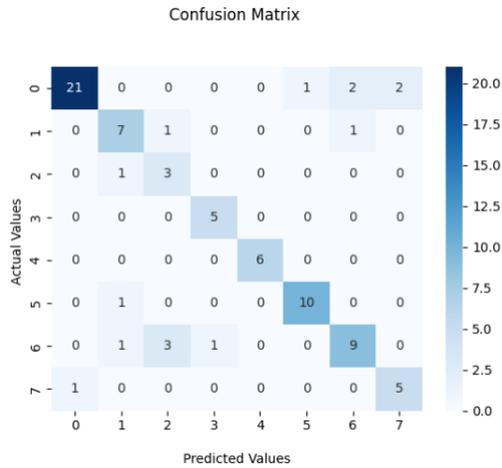
(a) Customized Tunned MobileNetV2

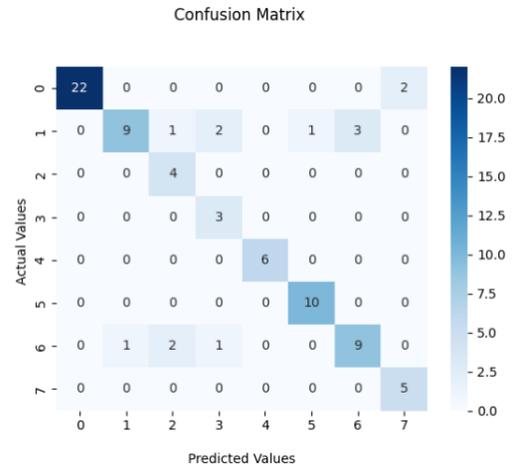
(b) Customized Tunned NASNetLarge

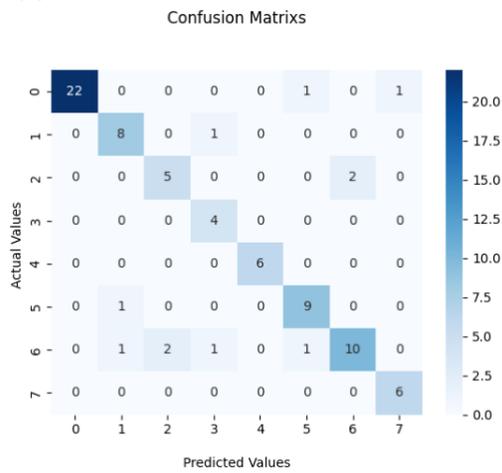
(c) Customized Tunned Xception.

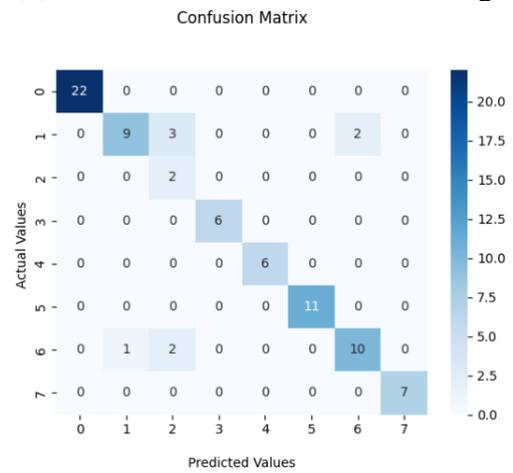
(d) Customized Tunned DenseNet201.

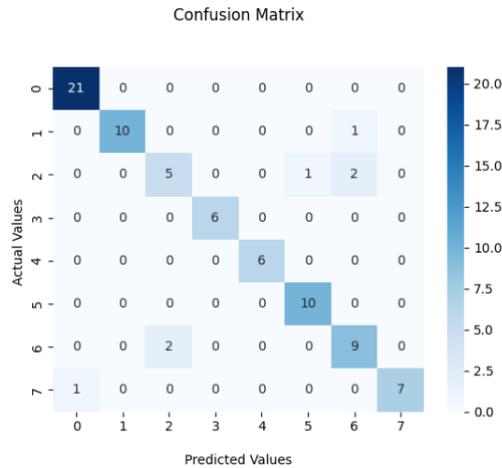
(e) Customized Tunned InceptionV3 (PotatoPestNet).

**Figure 8.** Confusion Matrix of the customised tunned models.

To comprehensively evaluate the models and assess their robustness in handling unbalanced classes, we utilized classification metrics such as precision, recall, F1-score, and accuracy. These metrics were calculated for each class individually, resulting in eight precision, recall, and F1-score values for each model. To present a consolidated performance assessment, we employed both macro average and weighted average methods. The macro average treats all classes equally, while the weighted average accounts for class imbalances by assigning more weight to classes with a greater number of instances. The aggregated metrics for all models are presented in Table 12. CTInceptionV3 (PotatoPestNet) outperformed the other models across all metrics, as indicated by the highlighted and underlined values in Table 12.

**Table 12.** Performance evaluation of the models.

| Types of Average | Model | Precision | Recall | F1-core | Accuracy |
|---|---|---|---|---|---|
| Macro Average | CTMobileNetV2 | 79 | 84 | 80 | 81 |
| | CTNASNetLarge | 79 | 90 | 82 | `84 |
| | CTXception | 84 | 89 | 85 | 86 |
| | CTDeseNet201 | 88 | 93 | 87 | 90 |
| | **CTInceptionV3 (PotatoPestNet)** | **92** | **90** | **91** | **91** |
| Weighted average | CTMobileNetV2 | 84 | 81 | 82 | 81 |
| | CTNASNetLarge | 87 | 84 | 84 | 84 |
| | CTXception | 87 | 86 | 86 | 86 |
| | CTDeseNet201 | 91 | 90 | 91 | 90 |
| | **CTInceptionV3 (PotatoPestNet)** | **91** | **91** | **91** | **91** |

To further evaluate the models' performance, we utilized receiver operating characteristic (ROC) curves, as shown in Figure 9. These curves illustrate the true positive rates against the false positive rates for all classes. ROC curves provide a comprehensive assessment of the models' classification performance across different thresholds. Ideally, a model's performance is considered favorable when the curve is closer to the point (1, 0), indicating higher true positive rates and lower false positive rates.

Based on our analysis, CTInceptionV3 (PotatoPestNet) demonstrated the highest classification performance among the considered models. Its ROC curve consistently exhibited superior performance across all classes, indicating its ability to accurately differentiate between pest and non-pest instances. The implementation on Gradio API is shown in Figure 10.

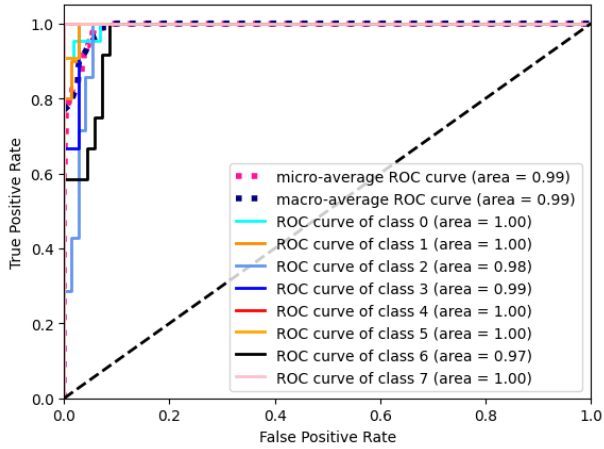
(a) Customized Tunned DenseNet201.

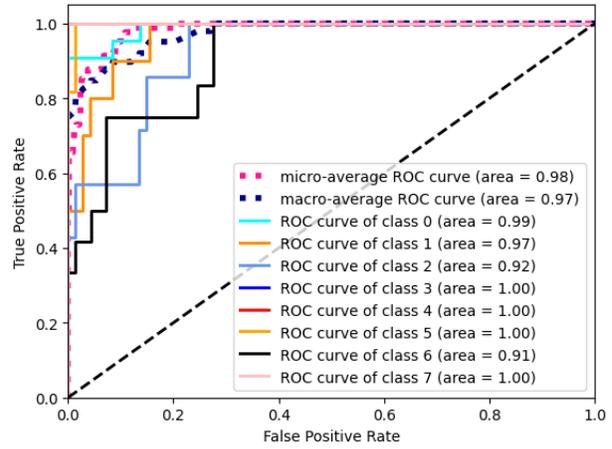
(b) Customized Tunned MobileNetV2.

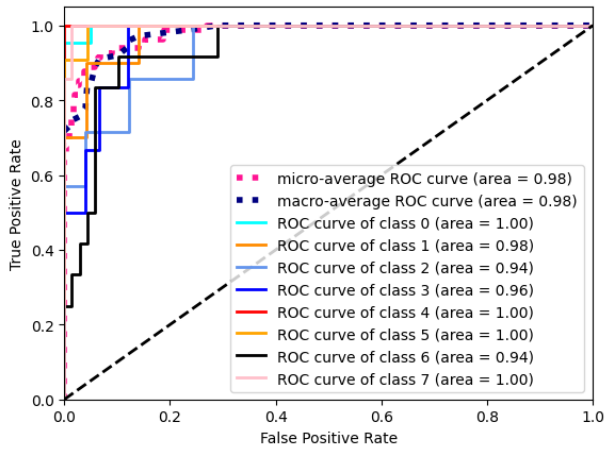
(c) Customized Tunned NASNetLarge.

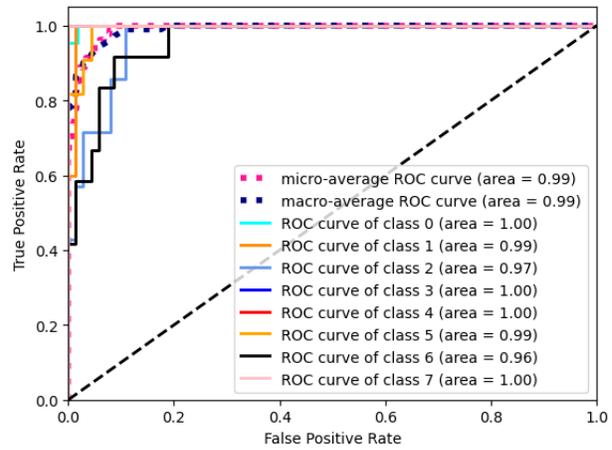
(d) Customized Tunned Xception.

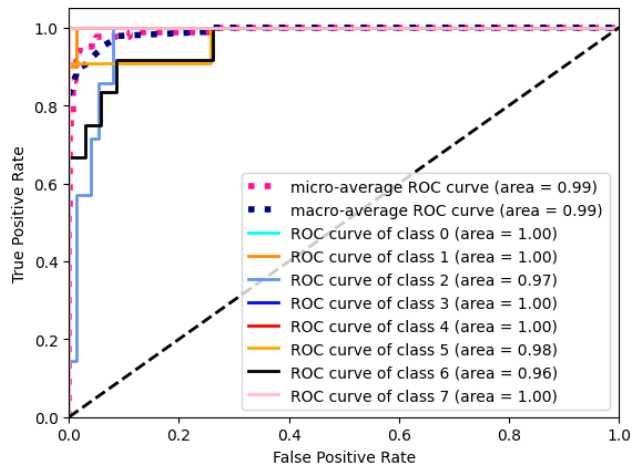
(e) Customized Tunned InceptionV3 (PotatoPestNet).

**Figure 9.** ROC curve of the customized tuned models.

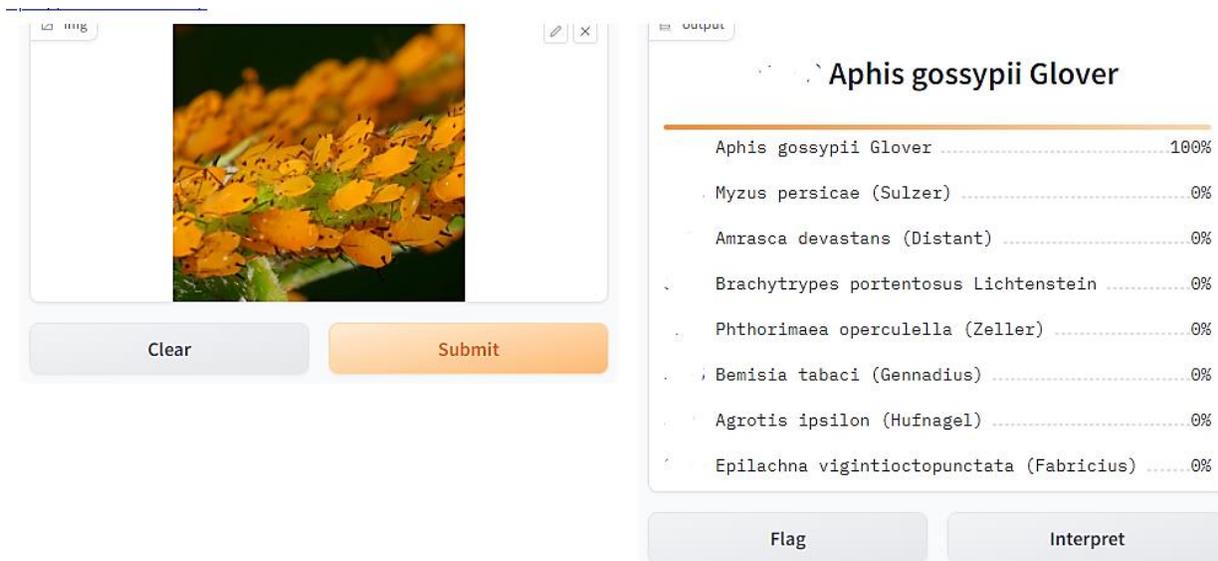

**Figure 10.** Visual evaluation in gradio API.

## 6. Discussion

In the result analysis, it is clearly shown that the customized tuned pre-trained models performed well. Since R.Sujatha et al. [13] concluded that CNN performed better than traditional ML, On the other hand, large training parameters as well as more computational time are in ANN [14], so we focused on the CNN. But the main challenge was the limited size of our dataset. That's why we changed our research methodology to move to deep transfer learning based on the research outcome of C. Jackulin et al. [15]. Another problem named overfitting appeared in transfer learning. We started to solve this issue and modified the most prominent of the five pre-trained models. The replacement of the fully connected layer with a global average pooling layer and the addition of a dropout layer were the magical solutions in our work. But at that time, the question was how much dropout rate, how much learning rate, and what kinds of optimizers we would use. We applied the random search technique with 10 trials and 20 epochs in each trial to find the best combination of targeting to get the best validation accuracy. The results in Tables 6–10 is a great reflection of different combinations of the parameters. The best parameters summarized in Table 11 are used in our modified models, which is another remarkable solution to avoiding overfitting and getting better performance. The loss curve, accuracy cure in Figure 7, and confusion matrix in Figure 8 presented the performances of the models at the learning stage and testing phase. All the models are optimally fitted to the loss curve and accuracy curve. There is no overfitting. The difference between validation loss and training loss of CTInceptionV3 (PotatoPestNet) is (0.4930-0.1922) = 0.3008, which is minimal, and the difference between training accuracy and validation accuracy is (0.9621-0.9155) = 0.0466, which is also minimal among the other models. It indicates that the PotatoPestNet model learned very well. From this point on, we paid extra attention to the testing phase of this model. The models obviously outperformed the other models. In the confusion matrix, the number of exact classifications by this model is 74, which is the highest accurate classification value. Table 12 also explores its superior performance to the others. The precision, recall, f1 score, and accuracy in macro average are 90%, 90%, 91%, and 91%, which are well balanced and highest in terms of values. Since our dataset is unbalanced, we also cross-checked the weighted average precision, recall, f1 score, and accuracy, which have the same value of 91%. That means our proposed PotatoPestNet models worked well.

In our work, the ROC curve in Figure 9 provided valuable insights into the performance analysis and robustness of the unbalanced dataset. We observed that all models exhibited curves that approached the point (1,0), indicating excellent classification performance. Notably, CTInceptionV3 stood out as the best-performing model, with its ROC curve consistently positioned closer to the upper-left corner. This suggests a higher TPR and lower FPR, indicating a superior ability to accurately classify positive instances while minimizing false positives. These findings align with the higher area under the ROC curve (AUC) value achieved by CTInceptionV3, further validating its superior overall performance. The ROC curve analysis has provided valuable insights for selecting the most suitable model and has implications for practical use, as it enables informed decisions on selecting an optimal classification threshold based on the relative costs of false positives and false negatives.

## 7. Limitations

Limitations are very common parts of any research. Similar to this research, this one also faced some limitations while conducting the experiments, which are discussed below.

- **Limited Dataset:** For training, validation, and testing, pre-trained models require a big and diverse dataset. We used a small size of dataset that's leads out model to be overfitted and narrows the scope of diversity of dataset.
- Only eight types of potato pests have been considered in our work.
- Different phase of the life cycle of pests are not taken into account.
- Only five pre-trained models are examined in our research.
- **Inadequate hardware**: Running pre-trained models demands substantial processing resources, such as high-end CPUs and GPUs. If the hardware is inadequate, it may take longer for the models to process, or the results may be impacted.

## 8. Conclusion and Future work

This study is intended to identify and classify potato pests. This experiment began with an identification of relevant research publications, a comparison of how other researchers recognized these types of issues, and an analysis of their methods for identifying and classifying relevant problems. This study devised its own methodology in which five CT-pre-trained machine learning models were used to classify potato pests and an efficient, robust PotatoPestNet model was proposed. The dataset was prepared very carefully and with the help of experts. Modification of the pre-trained models by replacing the fully connected layer with a global average pooling layer, adding a dropout layer, and finally tuning with a random search technique was the ace in the hole in potato pest detection. The CTInceptionV3-RS-Based approach considerably outperforms in terms of classification accuracy, precision, recall, and F1-score at 91% and saves training time compared to training from scratch, as demonstrated in the test results. This study has enormous implications for the agriculture business, as accurate and efficient identification of pests enables farmers to take prompt and targeted measures to prevent or control infestations. Our proposed PotatoPestNet model can also eliminate the need for costly and time-consuming data gathering and annotation during training.

In the future, this research could be investigated in more detail by adding the other 11 types of pests. Other deep learning approaches and techniques can be applied to enhance the precision of pest recognition models. In addition, applying these models to real-world settings and evaluating their effectiveness under varied climatic conditions might be a fruitful avenue for future research. Overall, our study has been successfully implemented, and we hope it paves the way for future research and activities by farmers.

**Author contribution:** Md. Simul Hasan Talukder proposed the machine learning concept and implemented the whole research work in Python environment. He wrote the materials, methodology, result analysis, and discussion sections. Rejwan Bin Sulaiman contributed to collecting the dataset, making ideas, and guiding the whole team. Mohammad Raziuddin Chowdhury greatly contributed to the preparation of the dataset. Musarrat Saberin Nipun wrote and analyzed the literature review. She also assisted in dataset preparation. Taminul Islam contributed in drafting and reviewing the manuscript.

**Declaration of Competing Interest:** The authors declare no conflict of interest.

**Dataset:** The dataset is available upon request.